\pdfoutput=1
\documentclass[11pt,a4paper]{article}
\usepackage[hyperindex,breaklinks]{hyperref}
\usepackage[hyperref]{acl2021}
\usepackage{amsmath}
\usepackage{times}
\usepackage{latexsym}
\usepackage{float}
\usepackage{mathtools}
\restylefloat{table}
\usepackage{xurl}

\usepackage{microtype}

\aclfinalcopy

\usepackage{cleveref}

\title{Cascaded Span Extraction and Response Generation\\for Document-Grounded Dialog}

\author{
  Nico Daheim,\textsuperscript{\rm 1}
  David Thulke,\textsuperscript{\rm 1,2}
  Christian Dugast,\textsuperscript{\rm 1,2}
  Hermann Ney\textsuperscript{\rm 1,2}
\\
\textsuperscript{\rm 1} Human Language Technology and Pattern Recognition Group, RWTH Aachen University, Germany \\
\textsuperscript{\rm 2} AppTek GmbH, Aachen, Germany \\
\{daheim, thulke, dugast, ney\}@i6.informatik.rwth-aachen.de
}

\date{}

\begin{document}
\maketitle
\begin{abstract}
  This paper summarizes our entries to both subtasks of the first DialDoc shared task which focuses on the agent response prediction task in goal-oriented document-grounded dialogs.
  The task is split into two subtasks: predicting a span in a document that grounds an agent turn and generating an agent response based on a dialog and grounding document.
  In the first subtask, we restrict the set of valid spans to the ones defined in the dataset, use a biaffine classifier to model spans, and finally use an ensemble of different models.
  For the second subtask, we use a cascaded model which grounds the response prediction on the predicted span instead of the full document.
  With these approaches, we obtain significant improvements in both subtasks compared to the baseline.
\end{abstract}

\section{Introduction}

Unstructured documents contain a vast amount of knowledge that can be useful information for responding to users in goal-oriented dialog systems.
The shared task at the first DialDoc Workshop focuses on grounding and generating agent responses in such systems.
Therefore, two subtasks are proposed: given a dialog extract the relevant information for the next agent turn from a document and generate a natural language agent response based on dialog context and grounding document.
In this paper, we present our submissions to both subtasks.

In the first subtask, we focus on modeling spans directly using a biaffine classifier and restricting the model's output to valid spans.
We notice that replacing BERT with alternative language models results in significant improvements.
For the second subtask, we notice that providing a generation model with an entire, possibly long, grounding document often leads to models struggling to generate factually correct output.
Hence, we split the task into two subsequent stages, where first a grounding span is selected according to our method for the first subtask which is then provided for generation.
With these approaches, we report strong improvements over the baseline in both subtasks.
Additionally, we experimented with marginalizing over all spans in order to be able to account for the uncertainty of the span selection model during generation.

\section{Related Work}
Recently, multiple datasets and challenges concerning conversational question answering have been proposed.
For example, \citet{saeidi2018} introduced ShARC, a dataset containing ca.\ 32k utterances which include follow-up questions on user requests which can not be answered directly based on the given dialog and grounding.
Similarly, the CoQA dataset \cite{reddy2019} provides 127k questions with answers and grounding obtained from human conversations.
Closer related to the DialDoc shared task, the task in the first track of DSTC~9 \cite{kim2020} was to generate agent responses based on relevant knowledge in task-oriented dialog.
However, the considered knowledge has the form of FAQ documents, where snippets are much shorter than those considered in this work.

Pre-trained trained language models such as BART \cite{lewis2020bart} or RoBERTa \cite{liu2019} have recently become a successful tool for different kinds of natural language understanding tasks, such as question answering (QA), where they obtain state-of-the-art results \cite{liu2019, clarkELECTRAPreTrainingText2020}.
Naturally, they have recently also found their way into task-oriented dialog systems \cite{lewis2020bart}, where they are either used as end-to-end systems \cite{budzianowskiHelloItGPT22019, hamEndtoEndNeuralPipeline2020} or as components for a specific subtask \cite{he2021learning}.

\section{Task Description}
The task of dialog systems is to generate an appropriate systems response $u_{T+1}$ to a user turn $u_T$ and preceding dialog context $u_1^{T-1} \coloneqq u_1, ..., u_{T-1}$.
In a document-grounded setting, $u_{T+1}$ is based on knowledge from a set of relevant documents $D^\prime \subseteq D$, where $D$ denotes all knowledge documents.
\citet{feng2020doc2dial} identify three tasks relevant to such systems, namely 1) user utterance understanding; 2) agent response prediction; 3) relevant document identification.
The shared task deals with the second task and assumes the result of the third task to be known.
They further split this task into \emph{agent response grounding prediction} and \emph{agent response generation}.
More specifically, one subtask focuses on identifying the grounding of $u_{T+1}$ and the second subtask on generating $u_{T+1}$.
In both subtasks exactly one document $d \in D$ is given.
Each document consists of multiple sections, whereby each section consists of a title and the content.
In the doc2dial dataset, the latter is split into multiple subspans.
In the following, we refer to these given subspans as \emph{phrases} in order to avoid confusing them with arbitrary spans in the document.

\paragraph{Agent Response Grounding Prediction}
\label{sec:response_grounding}
The first subtask is to identify a span in a given document that grounds the agent response \(u_{T+1}\).
It is formulated as a span selection task where the aim is
to return a tuple $(a_s, a_e)$ of start and end position of the relevant span within the grounding document $d$ based on the dialog history $u_1^T$.
In the context of the challenge, these spans always correspond to one of the given phrases in the documents.

\paragraph{Agent Response Generation}
The goal of response generation is to provide the user with a system response $u_{T+1}$ that is based on the dialog context $u_1^T$ and document $d$ and fits naturally into the preceding dialog.

\section{Methods}

\subsection{Baselines}
\paragraph{Agent Response Grounding Prediction}
For the first subtask, \citet{feng2020doc2dial} fine-tune BERT for question answering as proposed by \citet{devlin2019bert}.
Therefore, a start and end score for each token is calculated by a linear projection from the last hidden states of the model.
These scores are normalized using a softmax over all tokens to obtain probabilities for the start and end positions.
In order to obtain the probability of a specific span, the probabilities of the start and end positions are multiplied.
If the length of the documents exceeds the maximum length supported by the model, a sliding window with stride over the document is used and each window is passed to the model.
In training, if the correct span is not included in the window, the span only consisting of the begin of sequence token is used as target.
In decoding the scores of all windows are combined to find the best span.

\paragraph{Agent Response Generation} The baseline provided for the shared task uses a pre-trained BART model \cite{lewis2020bart} to generate agent responses.
The model is fine-tuned on the tasks training data by minimizing the cross-entropy of the reference tokens.
As input, it is provided with the dialog context, title of the document, and the grounding document separated by special tokens.
Inputs longer than the maximum sequence length supported by the model (1,024 tokens for BART) are truncated.
Effectively, this means that parts of the document are removed that may include the information relevant to the response.
An alternative to truncating the document would be to truncate the dialog context (i.e. removing the oldest turns which may be less relevant than the document).
We did not do experiments with this approach in this work and always included the full dialog context in the input.
For decoding beam search with a beam size of 4 is used.

\subsection{Agent Response Grounding Prediction}

\paragraph{Phrase restriction}
In contrast to standard QA tasks, in this task, possible start and end positions of spans are restricted to phrases in the document.
This motivated us to also restrict the possible outputs of the model to these positions.
That is, instead of applying the softmax over all tokens, it is only applied over tokens corresponding to the start or end positions of a phrase 
and thus only consider these positions in training and decoding.

\paragraph{Span-based objective}
The training objective for QA assumes that the probability of the start and end position are conditionally independent.
Previous work \cite{fajcikRethinkingObjectivesExtractive2020} indicates that directly modeling the joint probability of start and end position can improve performance.
Hence, to model this joint probability, we use a biaffine classifier as proposed by \citet{dozatDeepBiaffineAttention2017} for dependency parsing.

\paragraph{Ensembling} In our submission, we use an ensemble of multiple models for the prediction of spans to capture their uncertainty.
More precisely, we use Bayesian Model Averaging \cite{bma}, where the probability of a span $a = (a_s, a_e)$ is obtained by marginalizing the joint probability of span and model over all models $H$ as:
\begin{align}
  p\left(a \mid u_1^T, d\right) &= \sum_{h \in H} p_h\left(a \mid u_1^T, d\right) \cdot p\left(h\right)
\end{align}
The model prior $p\left(h\right)$ is obtained by applying a softmax function over the logarithm of the F1 scores obtained on a validation set.
Furthermore, we approximate the span posterior distribution $p_h\left(a \mid u_1^T, d\right)$ by an n-best list of size 20.

\subsection{Agent Response Generation}
\paragraph{Cascaded Response Generation}
One main issue with the baseline approach is that the model appears to be unable to identify the relevant knowledge when provided with long documents.
Additionally, due to the truncation, the input of the model may not even contain the relevant parts of the document.
To solve this issue, we propose to model the problem by cascading span selection and response generation.
This way, we only have to provide the comparatively short grounding span to the model instead of the full document.
This allows the model to focus on generating an appropriate utterance and less on identifying relevant grounding information.

Similar to the baseline, we fine-tune BART \cite{lewis2020bart}.
In training, we provide the model with the dialog context $u_1^T$ concatenated with the document title and reference span, each separated by a special token.
In decoding, the reference span is not available and we use the span predicted by our span selection model as input.

\paragraph{Marginalization over Spans}
Conditioning on only the ground truth span creates a mismatch between training and inference time since the ground truth span is not available at test time but has to be predicted.
This leads to errors occurring in span selection being propagated in response generation.
Further, the generation model is unable to take the uncertainty of the span selection model into account.
Similar to \citet{lewisRetrievalAugmentedGenerationKnowledgeIntensive2020} and \citet{Thulke2021rag} we propose to marginalize over all spans $S$.
We model the response generation as:
\begin{align*}
  p\left( \hat{u} = u_{T+1} \mid u_1^T; d \right) &= \\
  \prod_i^N \sum_{s \in S} \: & p\left(\hat{u}_i, s \mid \hat{u}_1^{i-1}; u_1^T; d \right)
\end{align*}
where the joint probability may be factorized into a span selection model $p\left( s \mid u_1^T; d \right)$ and a generation model $p\left( u_{T+1} \mid u_1^T,s; d \right)$ corresponding to our models for each subtask.
For efficiency, we approximate $S$ by the top 5 spans which we renormalize to maintain a probability distribution.
The generation model is then trained with cross-entropy using an n-best list obtained from the separately trained selection model.
A potential extension which we did not yet try is to train both models jointly.

\begin{table*}[h]
  \centering
  \caption{Results of our best system on test and validation set.}
  \label{table:main_results}
  \begin{tabular}{|l|r|r|r|r|l|r|r|}
    \hline
    \multicolumn{5}{|l|}{Subtask 1} & \multicolumn{3}{l|}{Subtask 2} \\
     & \multicolumn{2}{|r|}{test} & \multicolumn{2}{|r|}{val} & & test & val \\
    \hline
    model & F1 & EM & F1 & EM & model & \multicolumn{2}{|r|}{BLEU} \\ \hline\hline
    baseline & 67.9 & 51.5 & 70.8 & 56.3 & baseline (ours) & 28.1 & 32.9 \\
    RoBERTa & 73.2 & 58.3 & 77.3 & 65.6 & cascaded (RoBERTa) & 39.1 & 39.6 \\
    ensemble & \textbf{75.9} & \textbf{63.5} & \textbf{78.8} & \textbf{68.4} & cascaded (ensemble) & \textbf{40.4} & \textbf{41.5} \\
    \hline
  \end{tabular}
\end{table*}

\section{Data}
The shared task uses the doc2dial dataset \cite{feng2020doc2dial} which contains 4,793 annotated dialogs based on a total of 487 documents.
All documents were obtained from public government service websites and stem from the four domains \textit{Social Security Administration (ssa)}, \textit{Department of Motor Vehicles (dmv)}, \textit{United States Department of Veterans Affairs (va)}, and \textit{Federal Student Aid (studentaid)}.
In the shared task, each document is associated with exactly one domain and is annotated with sections and phrases.
The latter is described by a start and end index within the document and associated with a specific section that has a title and text.
Each dialog is based on one document and contains a set of turns.
Turns are taken either by a \textit{user} or an \textit{agent} and described by a dialog act and a list of grounding reference phrases in the document.

The training set of the shared task contains 3,474 dialogs with in total 44,149 turns.
In addition to the training set, the shared task organizers provide a validation set with 661 dialogs and a testdev set with 198 dialogs which include around 30\% of the dialogs from the final test set.
The final test set includes an additional domain of unseen documents and comprises a total of 787 dialogs.
Documents are rather long, have a median length of 817.5, and an average length of 991 tokens (using the BART subword vocabulary).
Thus, in many cases, truncation of the input is required.

\section{Experiments}

We base our implementation\footnote{Our code is made available at \url{https://github.com/ndaheim/dialdoc-sharedtask-21}} on the provided baseline code of the shared task \footnote{Baseline code is available at \url{https://github.com/doc2dial/sharedtask-dialdoc2021}}. 
Furthermore, we use the workflow manager Sisyphus \cite{peterSisyphusWorkflowManager2018} to organize our experiments.

For the first subtask, we use the base and large variants of RoBERTa \cite{liu2019} and ELECTRA \cite{clarkELECTRAPreTrainingText2020} instead of BERT large uncased.
In the second subtask, we use BART base instead of the large variant, which was used in the baseline code, since even after reducing the batch size to one, we were not able to run the baseline with a maximum sequence length of 1024 on our Nvidia GTX 1080 Ti and RTX 2080 Ti GPUs due to memory constraints.
All models are fine-tuned with an initial learning rate of 3e-5.
Base variants are trained for 10 epochs and large variants for 5 epochs.

We include agent follow-up turns in our training data, i.e. such turns $u_t$ made by agents, where the preceding turn $u_{t-1}$ was already taken by the agent.
Similar to other agent turns, i.e. where the preceding turn was taken by the user, these turns are annotated with their grounding span and can be used as additional samples in both tasks.
In the baseline implementation, these are excluded from training and evaluation.
To maintain comparability, we do not include them in the validation or test data.

For evaluation, we use the same evaluation metrics as proposed in the baseline.
In the first subtask, exact match (EM), i.e. the percentage of exact matches between the predicted and reference span (after lowercasing and removing punctuation, articles, and whitespace) and the token-level F1 score is used.
The second subtask is evaluated using SacreBLEU \cite{post2018call}.

\begin{table}[bh!]
  \centering
  \caption{Ablation analysis of our systems for subtask 1 on the validation set. The best single model results are underlined.}
  \label{table:ablation_task1}
\begin{tabular}{|l|r|r|r|}
  \hline
  model & F1 & EM & EM@5 \\
  \hline\hline
  baseline (BERT large) & 70.8 & 56.3 & 68.2 \\
  ELECTRA large & 75.1 & 63.1 & 79.5 \\
  RoBERTa large & \underline{77.3} & \underline{65.6} & 82.1 \\
  \; -- phrase restriction & 77.0 & 65.1 & 79.7\\
  \; \hphantom{--} -- follow-up turns & 76.5 & 64.5 & 80.9 \\
  \; -- follow-up turns & 75.7 & 63.2 & 80.3\\
  RoBERTa base & 74.8 & 63.1 & 79.5 \\
  \; + span-based & 73.6 & 62.5 & \underline{83.0} \\
  ensemble & \textbf{78.8} & \textbf{68.4} & \textbf{85.0} \\
  \hline
\end{tabular}
\end{table}

\subsection{Results}

\Cref{table:main_results} summarizes our main results and submission to the shared task.
The first line shows the results obtained by reproducing the baseline provided by the organizers (using BART base for Subtask 2).
We note that these results differ from the ones reported in \citet{feng2020doc2dial} due to slightly different data conditions in the shared task and their paper.
The second line shows the results of our best single model.
In Subtask 1, we obtained our best results by using RoBERTa large, trained additionally on agent follow-up turns, and by restricting the model to phrases occurring in the document.
Using an ensemble of this model, an ELECTRA large model trained with the same approach, and a RoBERTa base model trained with the span-based objective, we achieve our best result.
In the second subtask, our cascaded approach using this model and BART base significantly outperforms the baseline by over 10\% absolute in BLEU.
Using the results of the ensemble in Subtask 2 also translates to a significant improvement in BLEU, which indicates a strong influence of the agent response grounding prediction task.

\subsection{Ablation Analysis}

\paragraph{Agent Response Grounding Prediction}

\Cref{table:ablation_task1} gives an overview of our ablation analysis for the first subtask.
In addition to F1 and EM, we report the EM@5 which we define as the percentage of turns where an exact match is part of the 5-best list predicted by the model.
This metric gives an indication of the quality of the n-best list produced by the model.
Both RoBERTa and ELECTRA large outperform BERT large concerning F1 and EM with RoBERTa large performing best.
Removing agent follow-up turns in training consistently degrades the results for both models.
Restricting the predictions of the model to valid phrases during training and evaluation gives consistent improvements in the EM and EM@5 scores.

Training RoBERTa base using the span-based objective, we observe degradations in F1 and EM but observe an improvement in EM@5 which indicates that it better models the distribution across phrases.
Due to instabilities during training, we were not able to train a large model with the span-based objective.
Additionally, we only did experiments with the biaffine classifier discussed in \Cref{sec:response_grounding}.
It would be interesting to compare the results with other span-based objectives as the ones proposed by \citet{fajcikRethinkingObjectivesExtractive2020}.

\paragraph{Agent Response Generation}
\begin{table}
  \centering
  \caption{Ablation analysis of our systems for subtask 2 on the validation set.}
  \label{table:ablation_task2}
\begin{tabular}{|l|r|}
  \hline
  model & BLEU \\
  \hline\hline
  baseline (ours) & 32.9 \\
  span marginalization & 38.4 \\
  cascaded (RoBERTa large) & \underline{39.6} \\
  \; + section title & 39.6 \\
  \; \hphantom{+} + extended context & 39.5 \\
  cascaded (ensemble) & 41.2 \\
  \; + follow-up turns & 41.2 \\
  \; \hphantom{+} + beam-size 6 & 41.3 \\
  \; \hphantom{+} \hphantom{+} + repetition-penalty & \textbf{41.5} \\
  \hline
  cascaded (ground truth) & 46.2 \\
  \hline
\end{tabular}
\end{table}
\Cref{table:ablation_task2} shows an ablation study of our results in response generation.
The results show that our cascaded approach outperforms the baseline by a large margin.
Further experiments with additional context, such as the title of a section or a window of 10 tokens to each side of the span, do not give improvements.
This indicates that the selected spans seem to be sufficient to generate suitable responses.
Furthermore, marginalizing over multiple spans leads to degradations, which might be because training is based on an n-best list from an uncertain model.
We observe our best results when using only the predicted span and a beam size of 6.
Furthermore, we add a repetition penalty of 1.2 \cite{keskar2019ctrl} to discourage repetitions in generated responses.

Finally, the last line of the table reports the results of the cascaded method when using ground truth spans instead of the spans predicted by a model.
That is, a perfect model for the first subtask would additionally improve the results by 4.7 points absolute in BLEU.

\section{Conclusion}

In this paper, we have described our submissions to both subtasks of the first DialDoc shared task.
In the first subtask, we have experimented with restricting the set of spans that can be predicted to valid phrases, which yields constant improvements in terms of EM.
Furthermore, we have employed a model to directly hypothesize entire spans and shown the benefits of combining multiple models using Bayesian Model Averaging.
In the second subtask, we have shown how cascading span selection and response generation improves results when compared to providing an entire document in generation.
We have compared marginalizing over spans to just using a single span for generation, with which we obtain our best results in the shared task.

\section*{Acknowledgements}
This work has received funding from the European Research Council (ERC) under the European Union's Horizon 2020 research and innovation programme (grant agreement No 694537, project ``SEQCLAS''). The work reflects only the authors' views and the European Research Council Executive Agency (ERCEA) is not responsible for any use that may be made of the information it contains.

\bibliographystyle{acl_natbib}
\bibliography{anthology,acl2021}

\begin{thebibliography}{19}
\expandafter\ifx\csname natexlab\endcsname\relax\def\natexlab#1{#1}\fi

\bibitem[{Budzianowski and Vuli{\'c}(2019)}]{budzianowskiHelloItGPT22019}
Pawe{\l} Budzianowski and Ivan Vuli{\'c}. 2019.
\newblock \href {https://doi.org/10.18653/v1/D19-5602} {Hello, {{It}}'s
  {{GPT}}-2 -- {{How Can I Help You}}? {{Towards}} the {{Use}} of {{Pretrained
  Language Models}} for {{Task}}-{{Oriented Dialogue Systems}}}.
\newblock In \emph{Proceedings of the 3rd {{Workshop}} on {{Neural Generation}}
  and {{Translation}}}, {Hong Kong, China}. {Association for Computational
  Linguistics}.

\bibitem[{Clark et~al.(2020)Clark, Luong, Le, and
  Manning}]{clarkELECTRAPreTrainingText2020}
Kevin Clark, Minh-Thang Luong, Quoc~V. Le, and Christopher~D. Manning. 2020.
\newblock \href {https://openreview.net/pdf?id=r1xMH1BtvB} {{{ELECTRA}}:
  {{Pre}}-{{Training Text Encoders}} as {{Discriminators Rather Than
  Generators}}}.
\newblock In \emph{{{ICLR}}}.

\bibitem[{Devlin et~al.(2019)Devlin, Chang, Lee, and
  Toutanova}]{devlin2019bert}
Jacob Devlin, Ming-Wei Chang, Kenton Lee, and Kristina Toutanova. 2019.
\newblock \href {https://doi.org/10.18653/v1/N19-1423} {{BERT}: Pre-training of
  deep bidirectional transformers for language understanding}.
\newblock In \emph{Proceedings of the 2019 Conference of the North {A}merican
  Chapter of the Association for Computational Linguistics: Human Language
  Technologies, Volume 1 (Long and Short Papers)}, pages 4171--4186,
  Minneapolis, Minnesota. Association for Computational Linguistics.

\bibitem[{Dozat and Manning(2017)}]{dozatDeepBiaffineAttention2017}
Timothy Dozat and Christopher~D Manning. 2017.
\newblock \href {https://arxiv.org/pdf/1611.01734.pdf} {Deep {{Biaffine
  Attention}} for {{Neural Dependency Parsing}}}.
\newblock In \emph{{{ICLR}}}.

\bibitem[{Fajcik et~al.(2020)Fajcik, Jon, Kesiraju, and
  Smrz}]{fajcikRethinkingObjectivesExtractive2020}
Martin Fajcik, Josef Jon, Santosh Kesiraju, and Pavel Smrz. 2020.
\newblock \href {http://arxiv.org/abs/2008.12804} {{Rethinking the Objectives
  of Extractive Question Answering}}.

\bibitem[{Feng et~al.(2020)Feng, Wan, Gunasekara, Patel, Joshi, and
  Lastras}]{feng2020doc2dial}
Song Feng, Hui Wan, Chulaka Gunasekara, Siva Patel, Sachindra Joshi, and Luis
  Lastras. 2020.
\newblock \href {https://doi.org/10.18653/v1/2020.emnlp-main.652} {doc2dial: A
  goal-oriented document-grounded dialogue dataset}.
\newblock In \emph{Proceedings of the 2020 Conference on Empirical Methods in
  Natural Language Processing (EMNLP)}, pages 8118--8128, Online. Association
  for Computational Linguistics.

\bibitem[{Ham et~al.(2020)Ham, Lee, Jang, and
  Kim}]{hamEndtoEndNeuralPipeline2020}
Donghoon Ham, Jeong-Gwan Lee, Youngsoo Jang, and Kee-Eung Kim. 2020.
\newblock \href {https://doi.org/10.18653/v1/2020.acl-main.54} {End-to-{{End
  Neural Pipeline}} for {{Goal}}-{{Oriented Dialogue Systems}} using
  {{GPT}}-2}.
\newblock In \emph{Proceedings of the 58th {{Annual Meeting}} of the
  {{Association}} for {{Computational Linguistics}}}, pages 583--592, {Online}.
  {Association for Computational Linguistics}.

\bibitem[{He et~al.(2021)He, Lu, Bao, Wang, Wu, Niu, and Wang}]{he2021learning}
Huang He, Hua Lu, Siqi Bao, Fan Wang, Hua Wu, Zhengyu Niu, and Haifeng Wang.
  2021.
\newblock \href {http://arxiv.org/abs/2102.02096} {Learning to select external
  knowledge with multi-scale negative sampling}.

\bibitem[{Hoeting et~al.(1999)Hoeting, Madigan, Raftery, and Volinsky}]{bma}
Jennifer~A. Hoeting, David Madigan, Adrian~E. Raftery, and Chris~T. Volinsky.
  1999.
\newblock \href {http://www.jstor.org/stable/2676803} {Bayesian model
  averaging: A tutorial}.
\newblock \emph{Statistical Science}, 14(4):382--401.

\bibitem[{Keskar et~al.(2019)Keskar, McCann, Varshney, Xiong, and
  Socher}]{keskar2019ctrl}
Nitish~Shirish Keskar, Bryan McCann, Lav~R. Varshney, Caiming Xiong, and
  Richard Socher. 2019.
\newblock \href {http://arxiv.org/abs/1909.05858} {{CTRL:} {A} conditional
  transformer language model for controllable generation}.

\bibitem[{Kim et~al.(2020)Kim, Eric, Gopalakrishnan, Hedayatnia, Liu, and
  Hakkani-Tur}]{kim2020}
Seokhwan Kim, Mihail Eric, Karthik Gopalakrishnan, Behnam Hedayatnia, Yang Liu,
  and Dilek Hakkani-Tur. 2020.
\newblock \href {https://www.aclweb.org/anthology/2020.sigdial-1.35} {Beyond
  domain {API}s: Task-oriented conversational modeling with unstructured
  knowledge access}.
\newblock In \emph{Proceedings of the 21th Annual Meeting of the Special
  Interest Group on Discourse and Dialogue}, pages 278--289, 1st virtual
  meeting. Association for Computational Linguistics.

\bibitem[{Lewis et~al.(2020{\natexlab{a}})Lewis, Liu, Goyal, Ghazvininejad,
  Mohamed, Levy, Stoyanov, and Zettlemoyer}]{lewis2020bart}
Mike Lewis, Yinhan Liu, Naman Goyal, Marjan Ghazvininejad, Abdelrahman Mohamed,
  Omer Levy, Veselin Stoyanov, and Luke Zettlemoyer. 2020{\natexlab{a}}.
\newblock \href {https://doi.org/10.18653/v1/2020.acl-main.703} {{BART}:
  Denoising sequence-to-sequence pre-training for natural language generation,
  translation, and comprehension}.
\newblock In \emph{Proceedings of the 58th Annual Meeting of the Association
  for Computational Linguistics}, pages 7871--7880, Online. Association for
  Computational Linguistics.

\bibitem[{Lewis et~al.(2020{\natexlab{b}})Lewis, Perez, Piktus, Petroni,
  Karpukhin, Goyal, K{\"u}ttler, Lewis, Yih, Rockt{\"a}schel, Riedel, and
  Kiela}]{lewisRetrievalAugmentedGenerationKnowledgeIntensive2020}
Patrick Lewis, Ethan Perez, Aleksandara Piktus, Fabio Petroni, Vladimir
  Karpukhin, Naman Goyal, Heinrich K{\"u}ttler, Mike Lewis, Wen-tau Yih, Tim
  Rockt{\"a}schel, Sebastian Riedel, and Douwe Kiela. 2020{\natexlab{b}}.
\newblock \href {https://arxiv.org/pdf/2005.11401.pdf} {Retrieval-{{Augmented
  Generation}} for {{Knowledge}}-{{Intensive NLP Tasks}}}.
\newblock In \emph{{{NeurIPS}}}.

\bibitem[{Liu et~al.(2019)Liu, Ott, Goyal, Du, Joshi, Chen, Levy, Lewis,
  Zettlemoyer, and Stoyanov}]{liu2019}
Yinhan Liu, Myle Ott, Naman Goyal, Jingfei Du, Mandar Joshi, Danqi Chen, Omer
  Levy, Mike Lewis, Luke Zettlemoyer, and Veselin Stoyanov. 2019.
\newblock \href {http://arxiv.org/abs/1907.11692} {Roberta: A robustly
  optimized bert pretraining approach}.

\bibitem[{Peter et~al.(2018)Peter, Beck, and
  Ney}]{peterSisyphusWorkflowManager2018}
Jan-Thorsten Peter, Eugen Beck, and Hermann Ney. 2018.
\newblock \href {https://doi.org/10.18653/v1/D18-2015} {Sisyphus, a {{Workflow
  Manager Designed}} for {{Machine Translation}} and {{Automatic Speech
  Recognition}}}.
\newblock In \emph{Proceedings of the 2018 {{Conference}} on {{Empirical
  Methods}} in {{Natural Language Processing}}: {{System Demonstrations}}},
  pages 84--89, {Brussels, Belgium}. {Association for Computational
  Linguistics}.

\bibitem[{Post(2018)}]{post2018call}
Matt Post. 2018.
\newblock \href {https://www.aclweb.org/anthology/W18-6319} {A call for clarity
  in reporting {BLEU} scores}.
\newblock In \emph{Proceedings of the Third Conference on Machine Translation:
  Research Papers}, pages 186--191, Belgium, Brussels. Association for
  Computational Linguistics.

\bibitem[{Reddy et~al.(2019)Reddy, Chen, and Manning}]{reddy2019}
Siva Reddy, Danqi Chen, and Christopher~D. Manning. 2019.
\newblock \href {https://doi.org/10.1162/tacl_a_00266} {{C}o{QA}: A
  conversational question answering challenge}.
\newblock \emph{Transactions of the Association for Computational Linguistics},
  7:249--266.

\bibitem[{Saeidi et~al.(2018)Saeidi, Bartolo, Lewis, Singh, Rockt{\"{a}}schel,
  Sheldon, Bouchard, and Riedel}]{saeidi2018}
Marzieh Saeidi, Max Bartolo, Patrick S.~H. Lewis, Sameer Singh, Tim
  Rockt{\"{a}}schel, Mike Sheldon, Guillaume Bouchard, and Sebastian Riedel.
  2018.
\newblock \href {https://doi.org/10.18653/v1/d18-1233} {Interpretation of
  natural language rules in conversational machine reading}.
\newblock In \emph{Proceedings of the 2018 Conference on Empirical Methods in
  Natural Language Processing, Brussels, Belgium, October 31 - November 4,
  2018}, pages 2087--2097. Association for Computational Linguistics.

\bibitem[{Thulke et~al.(2021)Thulke, Daheim, Dugast, and Ney}]{Thulke2021rag}
David Thulke, Nico Daheim, Christian Dugast, and Hermann Ney. 2021.
\newblock \href {http://arxiv.org/abs/2102.04643} {{E}fficient {R}etrieval
  {A}ugmented {G}eneration from {U}nstructured {K}nowledge for
  {T}ask-{O}riented {D}ialog}.
\newblock In \emph{AAAI-21 : 9th Dialog System Technology Challenge (DSTC-9)
  Workshop}. 9th Dialog System Technology Challenge Workshop, online, 8 Feb
  2021 - 9 Feb 2021.

\end{thebibliography}

\appendix

\end{document}